%% file: merge_two_tex.tex
\documentclass[10pt,twocolumn,letterpaper]{article}

\usepackage{subfiles}
\usepackage{titling}  

\usepackage{iccv}
\usepackage{times}
\usepackage{epsfig}
\usepackage{graphicx}
\usepackage{amsmath}
\usepackage{amssymb}

\usepackage{microtype}
\usepackage{subfigure}
\usepackage{booktabs} 
\usepackage{tabularx}
\usepackage{mathtools}
\usepackage{amsthm}
\usepackage{verbatim}
\usepackage{dsfont}  
\usepackage{algorithm}
\usepackage{algorithmic}
\usepackage{multirow}
\usepackage{pifont}
\usepackage[table]{xcolor}

\usepackage{bigfoot}

\usepackage[pagebackref=true,breaklinks=true,letterpaper=true,colorlinks,bookmarks=false]{hyperref}

\usepackage[capitalize,noabbrev]{cleveref}

\usepackage{setspace}

\newcommand{\znote}[1]{\textcolor{cyan}{\textbf{ZW: #1}}}
\newcommand{\rnote}[1]{\textcolor{red}{\textbf{RX: #1}}}
\newcommand{\revise}[1]{\textcolor{black}{#1}}
\iccvfinalcopy 

\def\iccvPaperID{3339} 

\ificcvfinal\fi

\newcommand\blfootnote[1]{%
	\begingroup
	\renewcommand\thefootnote{}\footnote{#1}%
	\addtocounter{footnote}{-1}%
	\endgroup
}

\begin{document}

\subfile{myfirstfile}

\subfile{mysecondfile}

\end{document}

%% file: myfirstfile.tex
		\title{CMDA: Cross-Modality Domain Adaptation for Nighttime \\ Semantic Segmentation}
		
		\author{Ruihao Xia$^{1}$~~~~Chaoqiang Zhao$^{1}$~~~~Meng Zheng$^{2}$~~~~Ziyan Wu$^{2}$~~~~Qiyu Sun$^{1}$~~~~Yang Tang$^{1}$\thanks{Corresponding author.}
		\and
		$^{1}$East China University of Science and Technology~~~~$^{2}$United Imaging Intelligence
		\and
		{\tt\small \{xia\_rho,~zhaocq,~qysun\}@mail.ecust.edu.cn,~\{meng.zheng,~ziyan.wu\}@uii-ai.com}
		\and
		{\tt\small yangtang@ecust.edu.cn}
	}
	

	\maketitle
	\ificcvfinal\thispagestyle{empty}\fi
	
	\begin{abstract}
		Most nighttime semantic segmentation studies are based on domain adaptation approaches and image input. However, limited by the low dynamic range of conventional cameras, images fail to capture structural details and boundary information in low-light conditions.
		Event cameras, as a new form of vision sensors, are complementary to conventional cameras with their high dynamic range. To this end, we propose a novel unsupervised Cross-Modality Domain Adaptation (CMDA) framework to leverage multi-modality (Images and Events) information for nighttime semantic segmentation, with only labels on daytime images. 
		In CMDA, we design the Image Motion-Extractor to extract motion information and the Image Content-Extractor to extract content information from images, in order to bridge the gap between different modalities (Images $\rightleftharpoons$ Events) and domains (Day $\rightleftharpoons$ Night). Besides, we introduce the first image-event nighttime semantic segmentation dataset. Extensive experiments on both the public image dataset and the proposed image-event dataset demonstrate the effectiveness of our proposed approach. \revise{We open-source our code, models, and dataset at \url{https://github.com/XiaRho/CMDA}.}
	\end{abstract}
	
	\section{Introduction}
	Semantic segmentation is a crucial aspect of computer vision, which is essential for many applications, such as autonomous driving~\cite{SSAutonomous1, SSAutonomous2}, \revise{robotics~\cite{SSRobotic,SSRobotic2,SSRobotic3}}, and surveillance~\cite{SSSurveillance}. While semantic segmentation of daytime scenes has made significant progress~\cite{SSDay1, SSDay2, SegFormer,SSDay3}, challenges remain unsolved for nighttime scenes due to the much-degraded image quality at night, as well as the lack of high-quality annotations.
	Most existing works~\cite{SSNight_CCDistill, SSNight_DANNet, SSNight_DANIA,SSNight_CDAda} employed unsupervised domain adaptation (UDA) for nighttime semantic segmentation to solve the label scarcity problem, which leverage labeled daytime images (Source Domain) and unlabeled nighttime images (Target Domain). However, the low dynamic range of conventional frame-based cameras results in poor image quality at night compared to daytime images, \textit{i.e.,} the decrease in color contrast and details results in a reduction of clarity in nighttime images. This impedes the effective discrimination of object boundaries. Thus, the performance of methods solely relying on nighttime images as input is limited. 
	
	\begin{figure}[t]
		\centering
		\includegraphics[scale=0.66]{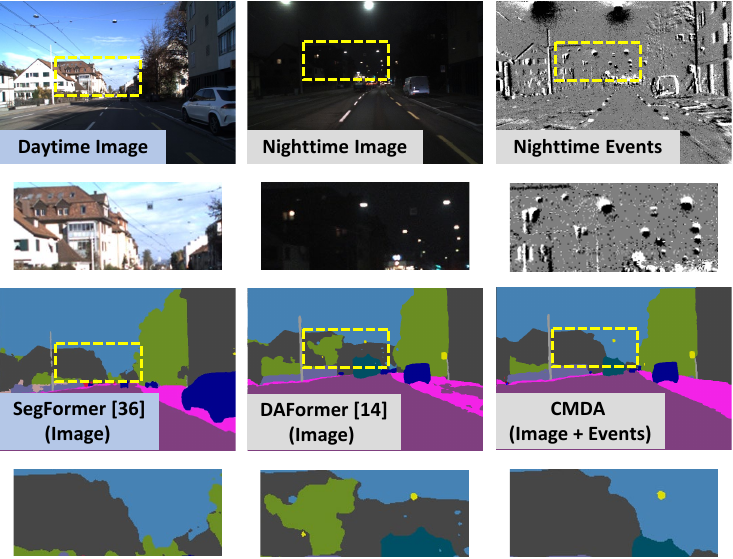}
		\caption{Images captured at different moments in the same location show that the low dynamic range of frame-based cameras leads to reduced color contrast and detailed edges of objects at night. To overcome this challenge, we introduce event cameras that have a high dynamic range and are capable of capturing more nighttime details. In comparison to the semantic segmentation results obtained from daytime images~\cite{SegFormer}, nighttime images result in misclassification cases~\cite{DAFormer}. However, our proposed CMDA improves this by introducing event modality for the first time.}
		\label{fig:idea}
	\end{figure}
	
	To address the limitations of frame-based cameras, we propose to employ event cameras for nighttime semantic segmentation. Event cameras output the spatio-temporal coordinates of pixels whose luminosity changes exceeding a certain threshold value~\cite{EventCameras1, EventCameras2}. Their unique operating principle offers a higher dynamic range (140 dB vs. 60 dB) over frame-based cameras~\cite{EventCamerasSurvey}, which enhances contrast in low-light scenarios, facilitating more precise segmentation of objects. On the other hand, events are asynchronous and spatially sparse, lacking a comprehensive representation of the scene. Hence methods based solely on events are typically inferior to image-based approaches~\cite{Dual, EvDistill}. To this end, we propose the first image-event cross-modality framework, Cross-Modality Domain Adaptation (CMDA), to leverage both image and event modalities for nighttime semantic segmentation in an unsupervised manner. As shown in Figure~\ref{fig:idea}, compared to conventional image-based UDA approaches, our framework achieves substantially improved nighttime semantic segmentation performance with the combination of event modality.
	
	In the proposed CMDA, the key challenges lie in establishing the connection between image and event modalities, as well as minimizing the domain shifts between the representations of daytime and nighttime images. Specifically:
	
	\textbf{Challenge 1: Images $\rightleftharpoons$ Events.} The absence of event modality in the source domain hinders the fusion of images and events. An intuitive idea is to transfer the daytime images into events. However, event cameras record the movement of the scene w.r.t. the camera, which cannot be determined with a single image. Thus, we propose the Image Motion-Extractor to extract motion information from adjacent images and bridge the gap between image and event modalities.
	
	\textbf{Challenge 2: Day $\rightleftharpoons$ Night.} Images can typically be separated into content and style information~\cite{SeparationStyleContent}. Previous image-based UDA approaches employed a style transfer network~\cite{CycleGAN} to transform daytime images so they look like nighttime~\cite{SSNight_CCDistill,SSNight_CDAda}. However, the transferred images are often unrealistic and unreliable, due to the significant and heterogeneous noise at night~\cite{StyleTransferDefect}. In contrast, we eliminate daytime and nighttime style information and preserve only content information based on the proposed Image Content-Extractor, which transfers both daytime and nighttime images to a common content domain.
	
	Then, we construct our network based on the image-based UDA method DAFormer~\cite{DAFormer}. Instead of taking only images as input, we combine events with images to perform improved nighttime semantic segmentation, with domain adaptation from labeled daytime images. In addition, as there are no existing benchmark datasets in the community for nighttime image-event semantic segmentation evaluation, we follow the image-based Dark Zurich dataset~\cite{DarkZurich} and manually annotate 150 image-event with fine, pixel-level labels from DSEC dataset~\cite{DSEC}.
	
	In summary, our contributions are as follows:
	
	\begin{itemize}
		\item 1) To the best of our knowledge, we introduce the first method to utilize event modality in nighttime semantic segmentation.
		\item 2) We propose a novel CMDA framework by fusing image and event modalities in an unsupervised manner with only labeled images from the source domain.
		\item 3) We propose the Image Motion-Extractor and Image Content-Extractor to bridge the gaps between modalities (Images $\rightleftharpoons$ Events) and domains (Day $\rightleftharpoons$ Night).
		\item 4) To fill in the missing evaluation criteria for nighttime image-event semantic segmentation, we align images and event modalities in the DSEC dataset~\cite{DSEC} and manually annotate 150 image-event with fine, pixel-level labels. The dataset and code will be made public.
		\item 5) We show the effectiveness of our CMDA framework, which achieves SOTA results on both the existing nighttime images benchmark dataset~\cite{DarkZurich} and our proposed image-event dataset.
	\end{itemize}

	\section{Related Work}
	\subsection{Event-based Semantic Segmentation}
	The problem of event-based semantic segmentation is under-explored, compared to image-based semantic segmentation due to the absence of high-quality datasets. Considering the paired image-event data in the DDD17 dataset~\cite{DDD17}, Alonso \itshape{et al.}\upshape~\cite{EvSeg} utilize a pretrained image-based network to generate pseudo labels for corresponding events. Then, labeled events data are employed to train an event-based network in a supervised manner.
	
	Considering the supervision on intermediate features, Wang \itshape{et al.}\upshape~\cite{EvDistill} utilize a pretrained image-based teacher network for cross-modality knowledge distillation. 
	Additionally, the training of the event-based network is aided by source data from another dataset~\cite{Cityscapes}. 
	Furthermore, Wang \itshape{et al.}\upshape~\cite{Dual} incorporate the cross-task knowledge transfer through an image reconstruction network to transfer the feature-level and prediction-level information. 
	Unlike previous studies, Sun \itshape{et al.}\upshape~\cite{ESS} employ a pretrained recurrent network, originally designed for image reconstruction~\cite{HighSpeed}, to encode events and generate semantic segmentation results. 
	However, the recurrent network requires a large number of events during both training and testing.
	
	\begin{figure*}
		\centering
		\centerline{\includegraphics[scale=0.75]{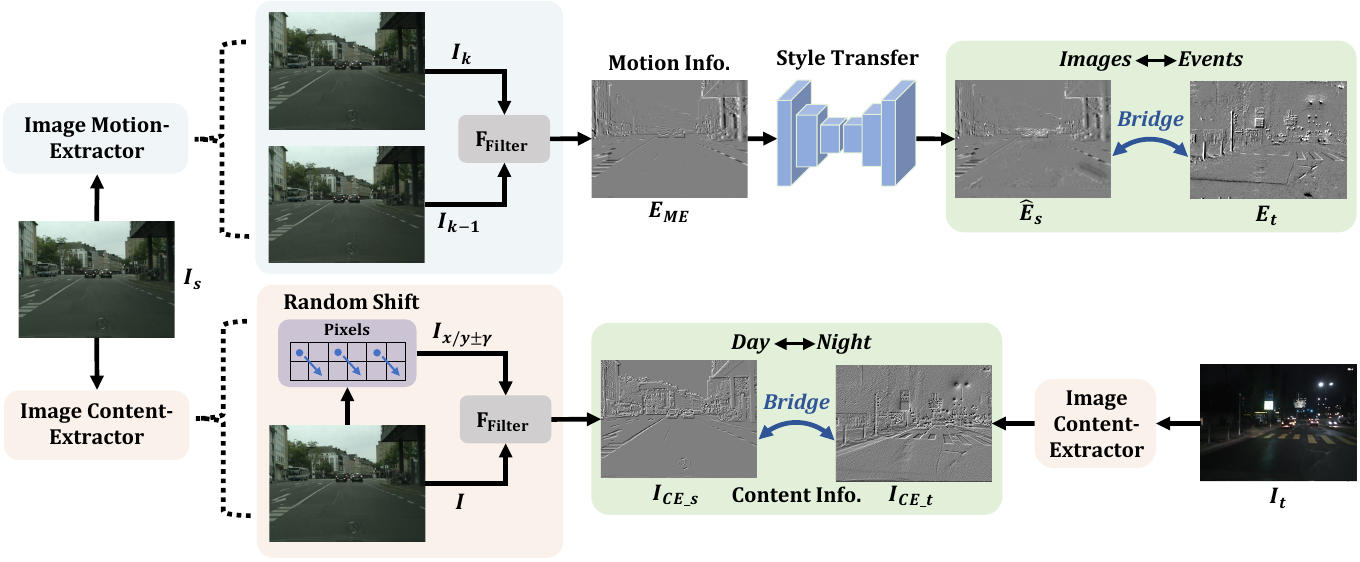}}
		\caption{Processed by Image Motion-Extractor and Image Content-Extractor, $E_{ME}$ and $I_{CE\_s/t}$ are utilized to bridge the gaps of different modalities (Images $I$ $\rightleftharpoons$ Events $E$) and domains (Source Daytime $s$ $\rightleftharpoons$ Target Nighttime $t$).}
		\label{fig:difference}
	\end{figure*}
	
	\revise{\textbf{Datasets.} Most of the existing event-based semantic segmentation datasets are synthetic datasets, \textit{e.g.,} EventScape~\cite{EventScape}, DELIVER~\cite{DELIVER}, and DADA-seg~\cite{DADASeg}. They are generated using simulators~\cite{CARLA} or pretrained networks~\cite{EventGAN}, resulting in large domain shifts compared with real-world events.}
	
	\revise{Other datasets like DDD17~\cite{DDD17} and DSEC~\cite{DSEC} record real-world events, but their semantic labels are generated by pretrained image-based networks~\cite{EvSeg,ESS} and only contain daytime scenes. Conversely for the first time, labels in nighttime scenes in our proposed DSEC Night-Semantic dataset are annotated manually.}

	\subsection{Nighttime Semantic Segmentation}
	\revise{Earlier approaches transfer daytime semantic knowledge to nighttime images via twilight images from different time periods~\cite{DarkModelAdaptation} or day-to-night style transfer networks~\cite{2019IV}.
		Then, the introduction of the paired day-night images dataset Dark Zurich~\cite{DarkZurich} propels advancements in this task.} 
	Sakaridis \itshape{et al.}\upshape~\cite{SSNight_MGCDA} transfer the labeled daytime dataset to twilight and night, utilizing curriculum learning to adapt to the unlabeled night domain. 
	Moving away from intermediate domains and models, Wu \itshape{et al.}\upshape~\cite{SSNight_DANNet,SSNight_DANIA} introduce an image relighting network and apply adversarial training. 
	Xu \itshape{et al.}\upshape~\cite{SSNight_CDAda} combine the inter-domain style adaptation and intra-domain gradual self-training to achieve smooth semantic knowledge transfer. 
	From the perspective of illumination and datasets differences, Gao \itshape{et al.}\upshape~\cite{SSNight_CCDistill} propose a novel domain adaptation framework via cross-domain correlation distillation. 
	However, paired day-night images are difficult to acquire in practical settings. 
	Recently, the emergence of transformer brings a huge boost to nighttime semantic segmentation, and our approach falls into this category. 
	These Transformer-based methods~\cite{DAFormer,MIC} employ self-training and consistency training to achieve superior performance without the need for paired data, which have achieved SOTA performance.
	
	However, day-to-night style transfer in Transformer-based methods leads to negative transfer, which is caused by the unrealistic and unreliable transferred nighttime images. Our proposed Image Content-Extractor transfers both domains to a shared content domain to alleviate the above issue. Then, we introduce event modality to make up for the low dynamic range of image modality for the first time.
	
	\section{Cross-Modality Domain Adaptation (CMDA)}
	In CMDA, given labeled images from the source domain $\left\{(I_s, Y_s)\right\}$ and unlabeled image-event pairs from the target domain $\left\{(I_t, E_t)\right\}$, our objective is to train a network $f$ that can accurately predict segmentation masks for the image-event pair input in the target domain, \textit{i.e.,} $f:(I_t, E_t)\rightarrow Y_t$. As there are no labels in the target domain, the key problem is to bridge the gaps between $I_s$ and $(I_t, E_t)$. Therefore, we design the Image Motion-Extractor to extract the motion information recorded by event cameras from $I_s$. Also, the Image Content-Extractor is designed to filter the style information and obtain the content information from both $I_s$ and $I_t$. In the following sections, we first introduce the key components of CMDA, \textit{i.e.,} the Image Motion-Extractor and Image Content-Extractor, followed by detailed explanations of CMDA structure as well as the training process.
	
	\begin{figure*}[t]
		\centering
		\includegraphics[scale=0.5]{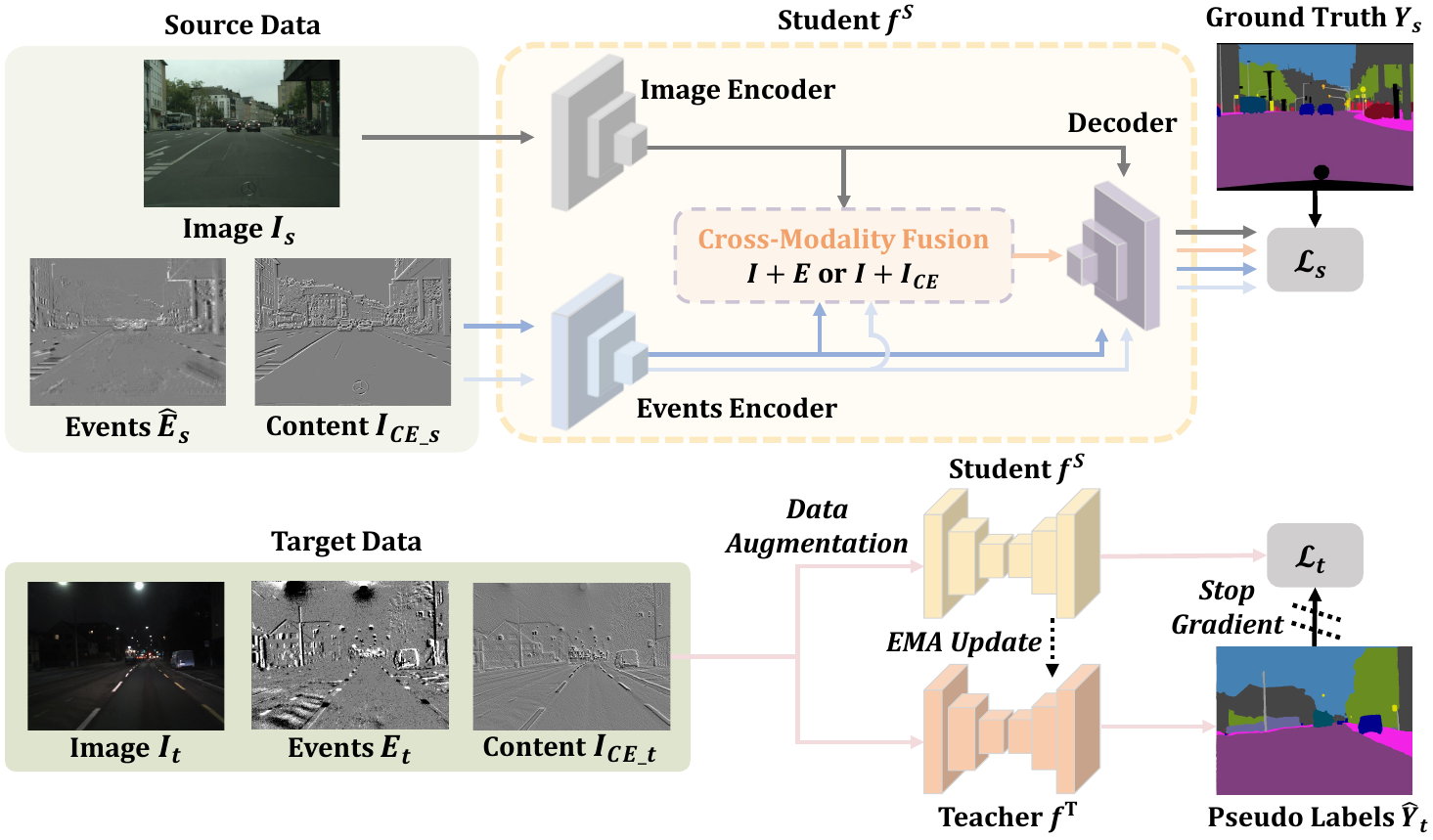}
		\caption{Two regularizations are employed to train the network: the supervised loss $\mathcal{L}_{s}$ in the source domain and the unsupervised domain adaptation loss $\mathcal{L}_{t}$ in the target domain. All losses are calculated on the student network $f^S$. The teacher network $f^T$ is used to generate pseudo labels for target data and updated with the EMA of $f^S$.}
		\label{fig:cmda}
	\end{figure*}
	
	\subsection{Image Motion-Extractor}
	The absence of event data in the source domain impedes the network to associate images with events. Considering that events are represented by the relative motion between the camera and the scene,  directly transferring images to events is non-trivial due to the lack of motion information in a single image. To overcome this challenge, we propose the Image Motion-Extractor to obtain the relative motion information $E_{ME}$ from two temporally adjacent images, as illustrated at the top of Figure~\ref{fig:difference}.
	
	Considering the event camera that records the logarithmic intensity change of pixels~\cite{EventCamerasSurvey}, we simulate this by differencing the same pixel of two adjacent images on the logarithmic domain. Thus, given by two temporally adjacent grayscale images $I_{k-1},I_k\in \mathbb{R} ^{H\times W\times 1}$, we compute $E_{ME}=\operatorname{F_{Filter}}(I_{k-1},I_k)$ with the following:
	\begin{align}
		&\operatorname{F_{Filter}}(I_1, I_2)=\operatorname{F_{Norm}}(\operatorname{F_{ClipIgn}}(\operatorname{F_{LogDiff}}(I_1, I_2))), \label{eqn:Extractor} \\
		&\operatorname{F_{LogDiff}}(I_1, I_2)=\ln{(I_1 + \epsilon)} - \ln{(I_2 + \epsilon)}, \label{eqn:LogDiff} \\
		&\operatorname{F_{ClipIgn}}(x)=\min{(|x|,\alpha)}\cdot \operatorname{sgn}(x) \cdot \mathds{1}(|x|>\beta), \label{eqn:ClipIgn} \\
		&\operatorname{F_{Norm}}(x)=2\cdot\frac{x-\min(x)}{\max(x)-\min(x)}-1,  \label{eqn:Norm}
	\end{align}
	where $\operatorname{F_{LogDiff}}(I_1, I_2)$ represents the difference of $I_1, I_2$ in the logarithmic domain, $\epsilon$ is a small scalar constant to prevent taking the logarithm of zero. $\operatorname{F_{ClipIgn}}(x)$ aims to clip larger values and ignore smaller values through two hyper-parameters $\alpha$ and $\beta$, $\mathds{1}(\cdot)$ is the indicator function, and $\operatorname{sgn}(\cdot)$ is the signum function. $\operatorname{F_{Norm}}(x)$ is the min-max normalization, scaling the values from -1 to 1.
	
	However, like frame-based cameras, event cameras are also suffering from noise at night. To further narrow the gap between $E_{ME}$ and $E_t$, we train a style transfer network~\cite{CycleGAN} $G_{E_{ME}\to E}$ in an unsupervised manner to add the style of $E_t$ to $E_{ME}$, resulting in transferred daytime events $\hat{E}_s=G_{E_{ME}\to E}(E_{ME})$. So far, we associate $I_s$ with $E_t$ with our proposed Image Motion-Extractor and $G_{E_{ME}\to E}$.
	
	\subsection{Image Content-Extractor}
	Previous image-based UDA approaches transferred daytime images $I_s$ to the nighttime style with a style transfer network~\cite{CycleGAN} to alleviate domain gaps~\cite{SSNight_CCDistill, SSNight_CDAda}. However, the real nighttime style is difficult to construct due to the complex and changing nighttime scenes~\cite{StyleTransferDefect}. Instead, we propose the Image Content-Extractor to obtain the content information. By eliminating the daytime and nighttime style, we transfer both $I_s$ and $I_t$ to the intermediate domain and discard the nighttime style generating and utilization of style transfer network.
	
	Given a grayscale image $I$, we shift it $\gamma$ pixels to the left/right and up/down randomly and obtain $I_{x \pm \gamma}$ and $I_{y \pm \gamma}$. Then, the intermediate shared content domain $I_{CE}$ is generated by the following:
	\begin{align}
		\label{eqn:ISF}
		I_{CE}=\frac{1}{2}\cdot \operatorname{F_{Filter}}(I, I_{x \pm \gamma})+\frac{1}{2}\cdot \operatorname{F_{Filter}}(I, I_{y \pm \gamma})
	\end{align}
	
	By subtracting the shifted version of the image from itself, pixels of the same color are erased, leaving only the pixels at the edges of the scene, \textit{i.e.,} content information.
	
	We process both $I_s$ and $I_t$ to obtain $I_{CE\_s}$ and $I_{CE\_t}$. As shown in Figure~\ref{fig:difference}, after converting $I$ into $I_{CE}$, the domain-specific texture (Style Information) is largely eliminated, and only the domain-invariant structure (Content Information) is retained. 
	
	\subsection{Network Details}
	The proposed extractors mentioned above enable us to bridge the gaps between modalities and domains at the input level. In this section, we elaborate on how to effectively utilize $I$, $E$ and $I_{CE}$ within the CMDA framework. 
	
	\textbf{Overview.} Our CMDA is based on the image-based self-training method DAFormer~\cite{DAFormer}. The framework comprises a student network $f^S$ and a teacher network $f^T$. Given source and target data as inputs, $f^S$ outputs predicted semantic segmentation results $P$. These results are then computed with the source ground truth and target pseudo labels to obtain the cross-entropy loss. $f^T$ aims to provide pseudo labels in the target domain and is updated with the exponentially moving average (EMA) of $f^S$.
	
	\textbf{Network Architecture.} As shown in Figure~\ref{fig:cmda}, both $f^S$ and $f^T$ consist of two encoders, \revise{one cross-modality fusion module}, and one decoder. Given $I/E/I_{CE}$, the image encoder extracts the features from $I$, while the events encoder extracts the features from both $E$ and $I_{CE}$. The fusion module is utilized to combine features from $I$ and $E/I_{CE}$. Finally, the decoder receives both the fused and non-fused features and generates predicted semantic segmentation outputs $P_I$, $P_E$, $P_{I_{CE}}$, and $P_{I+E}$/$P_{I+I_{CE}}$.
	
	\textbf{Fusion Module.} Both the image and events encoders in our framework generate features with four different scales. To fuse features from the same scale, we individually input them into the attention block adapted from SegFormer~\cite{SegFormer} and average them to obtain the fused features.
	
	\begin{algorithm}[t]
		\caption{Training of CMDA}
		\label{alg:CMDA}
		\begin{algorithmic}[1]
			\REQUIRE Source data $\left\{(I_s, Y_s)\right\}$, Target data $\left\{(I_t, E_t)\right\}$.
			\STATE Obtain $E_{ME}$, $I_{CE\_s}$, and $I_{CE\_t}$ based on Eqn. (\ref{eqn:Extractor}) and Eqn. (\ref{eqn:ISF}).
			\STATE Train $G_{E_{ME}\to E}$ and obtain $\hat{E}_s=G_{E_{ME}\to E}(E_{ME})$.
			\STATE \revise{Initialize $f^S$ and $f^T$ with the same pretrained network.}
			\FOR{$n=1$ {\bfseries to} $40k$}
			\STATE Compute source loss $\mathcal{L}_s$ based on Eqn. (\ref{eqn:source_target_loss}).
			\STATE \revise{Generate pseudo labels $\hat{Y}_t$ by randomly choosing $E$ or $I_{CE}$ to fuse with $I$.}
			\STATE Compute target loss $\mathcal{L}_t$ based on Eqn. (\ref{eqn:source_target_loss}).
			\STATE Loss back-propagation and update $f^S$.
			\STATE Update $f^T$ based on the EMA in Eqn. (\ref{eqn:ema}).
			\ENDFOR
		\end{algorithmic}
	\end{algorithm}
	
	\textbf{Random Choice of $E$ or $I_{CE}$.} \revise{To take full advantage of $E$ as well as $I_{CE}$ modalities, pseudo labels in the target domain are generated by fusing $I$ with $E$ or $I_{CE}$ randomly, \textit{i.e.,} $\hat{Y}_t=f^T(I_t, E_t/I_{CE\_t})$.}
	
	
	\textbf{Training Loss.} Given daytime modalities $I_s$, $\hat{E}_s$, $I_{CE\_s}$, and nighttime modalities $I_t$, $E_t$, $I_{CE\_t}$, we train the student network $f^S$ with a combination of several categorical cross-entropy (CE) losses $\mathcal{L}_{s/t}$ calculated with daytime ground truth $Y_s$ and nighttime pseudo labels $\hat{Y}_t$. For brevity, we omit the domain term $s/t$ of $P$ and $Y$ in the following:
	\begin{align}
		\label{eqn:source_target_loss}
		\mathcal{L}_{s/t}={} &\lambda_I\mathcal{L}_{ce}(P_{I}, Y) + \lambda_E\mathcal{L}_{ce}(P_{E}, Y) \notag \\
		& + \lambda_{I_{CE}}\mathcal{L}_{ce}(P_{I_{CE}}, Y) \notag \\ 
		& + \lambda_{Fusion}\mathcal{L}_{ce}(\revise{P_{I+E}}, Y), \\
		\mathcal{L}_{ce}(P,Y&)=\sum_{j=1}^{H\times W}\sum_{c=1}^{C}Y^{(j,c)}\log\delta(P^{(j,c)}),
	\end{align}
	where $\delta(P)$ denoted the softmax output of the predicted results $P$, $C$ is the number of semantic classes, $\lambda_I$, $\lambda_E$, $\lambda_{I_{CE}}$, and $\lambda_{Fusion}$ are hyper-parameters.
	
	In contrast to $f^S$, which is updated through gradient descent, $f^T$ is updated by the exponentially moving average (EMA) of the weights of $f^S$ in each training step following DAFormer~\cite{DAFormer}:
	\begin{align}
		\label{eqn:ema}
		f^T=\sigma f^T+(1-\sigma)f^S,
	\end{align}
	where $\sigma$ is a momentum parameter. 
	
	We summarize the overall training process of our CMDA framework in Algorithm \ref{alg:CMDA}.
	
	
	\begin{table}
		\begin{center}
			\begin{tabular}{@{}c@{\enspace}c@{\enspace}c@{}}
				\toprule
				Sequence & Training samples & Testing samples \\
				\midrule
				Zurich City 09a & 508 & 45 \\
				Zurich City 09b & 109 & 9 \\
				Zurich City 09c & 371 & 34 \\
				Zurich City 09d & 478 & 42 \\
				Zurich City 09e & 226 & 20 \\
				\textbf{Total} & \textbf{1,692} & \textbf{150}\\
				\bottomrule
			\end{tabular}
		\end{center}
		\caption{The dataset split of our proposed DSEC Night-Semantic dataset.}
		\label{tab:dataset}
	\end{table}
	
	\section{Experiments}
	
	\begin{figure*}[t]
		\centering
		\centerline{\includegraphics[scale=0.7]{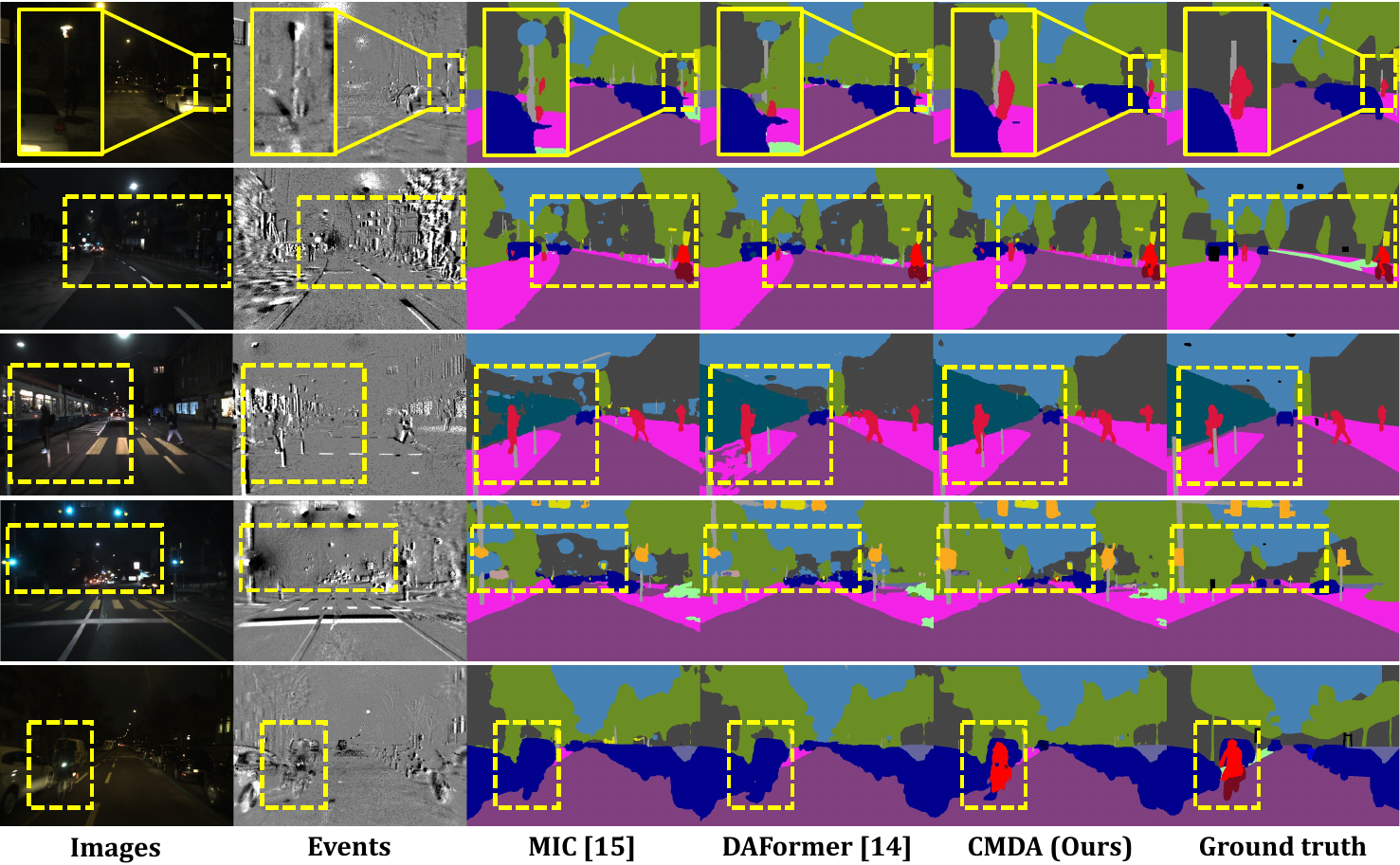}}
		\caption{Qualitative semantic segmentation results generated by image-based SOTA methods MIC~\cite{MIC}, DAFormer~\cite{DAFormer}, and our proposed CMDA in the DSEC Night-Semantic dataset.}
		\label{fig:STOA}
	\end{figure*}
	
	\begin{table*}[t]
		\begin{center}
			\begin{tabular}{@{}c@{\enspace}c@{\enspace}c@{\enspace}c@{\enspace}c@{\enspace}c@{\enspace}c@{\enspace}c@{\enspace}c@{\enspace}c@{\enspace}c@{\enspace}c@{\enspace}c@{\enspace}c@{\enspace}c@{\enspace}c@{\enspace}c@{\enspace}c@{\enspace}c@{\enspace}c@{}}
				\toprule
				Method & \rotatebox{90}{Road} & \rotatebox{90}{S.walk} & \rotatebox{90}{Build.} & \rotatebox{90}{Wall} & \rotatebox{90}{Fence} & \rotatebox{90}{Pole} & \rotatebox{90}{Tr.L.} & \rotatebox{90}{Tr.S.} & \rotatebox{90}{Veg.} & \rotatebox{90}{Terr.} & \rotatebox{90}{Sky} & \rotatebox{90}{Person} & \rotatebox{90}{Rider} & \rotatebox{90}{Car} & \rotatebox{90}{Bus} & \rotatebox{90}{Train} & \rotatebox{90}{M.bike} & \rotatebox{90}{Bike} & MIoU \\
				\midrule
				SePiCo$\dag$~\cite{SePiCo} & 93.3 & 58.7 & 56.8 & 28.2 & 4.7 & 34.1 & 27.9 & 55.1 & 55.7 & 56.1 & 76.1 & 50.5 & 30.5 & 75.1 & 75.5 & 71.0 & 22.6 & 26.6 & 49.9\\
				Refign$\dag$~\cite{Refign} & 92.2 & 56.6 & \bf{59.2} & 28.0 & 7.9 & 38.4 & 32.1 & \bf{60.0} & 56.9 & 57.5 & 79.6 & \bf{60.3} & 26.3 & 72.3 & 68.7 & 77.8 & 39.3 & 35.7 & 52.7  \\
				MIC~\cite{MIC} & 94.0 & 62.1 & 54.2 & 36.3 & \bf{9.8} & 37.7 & 29.2 & 48.4 & \bf{62.6} & 67.2 & 74.5 & 53.1 & 25.5 & 73.0 & 79.7 & 65.7 & 56.0 & 37.4 & 53.7  \\
				DAFormer~\cite{DAFormer} & 93.9 & 64.3 & 53.7 & 34.9 & 7.5 & 40.7 & 34.1 & 55.9 & 61.6 & 68.7 & 84.5 & 57.1 & 28.8 & 75.0 & 68.5 & 77.8 & 57.6 & 42.6 & 56.0\\
				\midrule
				Baseline($I$) & 94.2 & 64.5 & 44.8 & 36.3 & \bf{9.8} & 39.1 & 23.8 & 58.3 & 56.5 & 67.3 & 73.0 & 59.5 & 34.4 & 75.4 & 87.6 & \bf{78.8} & 42.6 & 45.2 & 55.1 \\
				CMDA($E$) & 90.8 & 50.9 & 59.1 & 30.5 & 4.4 & 26.2 & 28.1 & 41.6 & 53.5 & 49.6 & 68.3 & 33.9 & 30.2 & 68.0 & 65.5 & 57.3 & 41.9 & 28.6 & 46.0  \\
				CMDA($I$) & \bf{94.6} & 67.5 & 55.5 & 36.2 & 7.9 & 39.3 & 42.2 & 55.6 & 60.7 & 70.2 & \bf{85.4} & 50.7 & 39.3 & 77.6 & 84.8 & 73.9 & 53.2 & \bf{45.3} & 57.8 \\
				CMDA($I$+$E$) & \bf{94.6} & \bf{68.3} & 58.2 & \bf{37.5} & 8.8 & \bf{44.0} & \bf{45.7} & 57.7 & 61.4 & \bf{70.4} & 85.1 & 56.0 & \bf{45.9} & \bf{79.2} & \bf{87.8} & 73.8 & \bf{61.6} & 45.0 & \bf{60.1}\\
				\bottomrule
			\end{tabular}
		\end{center}
		\caption{Quantitative semantic segmentation results evaluated with MIoU (\%) in our proposed DSEC Night-Semantic Dataset. ($I$/$E$/$I$+$E$) indicates the input modalities during testing. The best result is highlighted in bold. $\dag$ denotes the methods utilizing additional coarsely aligned daytime images in the target domain which are not available in our dataset. We directly test their model trained on Dark Zurich~\cite{DarkZurich}.}
		\label{tab:STOA}
	\end{table*}
	
	\subsection{Implementation Detail}
	Our baseline model is adopted from DAFormer~\cite{DAFormer} without the loss of Thing-Class Feature Distance.
	Building upon this baseline, we incorporate an events encoder and a cross-modality fusion module into the network structure. For loss weighting, we use $\lambda_{I}=\lambda_{Fusion}=0.5$ and $\lambda_{E}=\lambda_{I_{CE}}=0.25$. For $E_{ME}$ and $I_{CE}$, we use $\alpha=0.1$, $\beta=0.005$, and $\gamma=1$ in Eqn. (\ref{eqn:ClipIgn}) and Eqn. (\ref{eqn:ISF}). 
	$E_t$ are selected within 50ms before the timestamps of $I_t$ and processed in the voxel grid representation~\cite{VoxelGrid}. It takes 40,000 iterations on a batch size of two to train our CMDA. All experiments are conducted on a Tesla A100 GPU.
	
	\begin{table*}[t]
		\begin{center}
			\begin{small}
				\begin{tabular}{@{}c@{\enspace}c@{\enspace}c@{\enspace}c@{\enspace}c@{\enspace}c@{\enspace}c@{\enspace}c@{\enspace}c@{\enspace}c@{\enspace}c@{\enspace}c@{\enspace}c@{\enspace}c@{\enspace}c@{\enspace}c@{\enspace}c@{\enspace}c@{\enspace}c@{\enspace}c@{\enspace}c@{}}
					\toprule
					Method & \rotatebox{90}{Road} & \rotatebox{90}{S.walk} & \rotatebox{90}{Build.} & \rotatebox{90}{Wall} & \rotatebox{90}{Fence} & \rotatebox{90}{Pole} & \rotatebox{90}{Tr.L.} & \rotatebox{90}{Tr.S.} & \rotatebox{90}{Veg.} & \rotatebox{90}{Terr.} & \rotatebox{90}{Sky} & \rotatebox{90}{Person} & \rotatebox{90}{Rider} & \rotatebox{90}{Car} & \rotatebox{90}{Truck} & \rotatebox{90}{Bus} & \rotatebox{90}{Train} & \rotatebox{90}{M.bike} & \rotatebox{90}{Bike} & MIoU \\
					\midrule
					MGCDA$\dag$~\cite{SSNight_MGCDA} & 80.3 & 49.3 & 66.2 & 7.8 & 11.0 & 41.4 & 38.9 & 39.0 & 64.1 & 18.0 & 55.8 & 52.1 & \bf{53.5} & 74.7 & \bf{66.0} & 0.0 & 37.5 & 29.1 & 22.7 & 42.5 \\ 
					DANNet$\dag$~\cite{SSNight_DANNet} & 90.0 & 54.0 & 74.8 & 41.0 & 21.1 & 25.0 & 26.8 & 30.2 & 72.0 & 26.2 & 84.0 & 47.0 & 33.9 & 68.2 & 19.0 & 0.3 & 66.4 & 38.3 & 23.6 & 44.3 \\ 
					CDAda$\dag$~\cite{SSNight_CDAda} & 90.5 & 60.6 & 67.9 & 37.0 & 19.3 & 42.9 & 36.4 & 35.3 & 66.9 & 24.4 & 79.8 & 45.4 & 42.9 & 70.8 & 51.7 & 0.0 & 29.7 & 27.7 & 26.2 & 45.0 \\
					DANIA$\dag$~\cite{SSNight_DANIA} & 90.8 & 59.7 & 73.7 & 39.9 & \bf{26.3} & 36.7 & 33.8 & 32.4 & 70.5 & 32.1 & 85.1 & 43.0 & 42.2 & 72.8 & 13.4 & 0.0 & 71.6 & 48.9 & 23.9 & 47.2 \\
					CCDistill$\dag$~\cite{SSNight_CCDistill} & 89.6 & 58.1 & 70.6 & 36.6 & 22.5 & 33.0 & 27.0 & 30.5 & 68.3 & 33.0 & 80.9 & 42.3 & 40.1 & 69.4 & 58.1 & 0.1 & 72.6 & 47.7 & 21.3 & 47.5 \\
					\revise{LoopDA$\dag$~\cite{SSNight_LoopDA}} & 92.1 & 63.3 & \bf{80.3} & 41.1 & 13.9 & 40.8 & 39.7 & 41.1 & 71.3 & 28.4 & 85.5 & 50.2 & 38.5 & 78.2 & 58.5 & 3.0 & 77.2 & 26.5 & 31.0 & 50.6 \\
					DAFormer~\cite{DAFormer} & 93.5 & 65.5 & 73.3 & 39.4 & 19.2 & 53.3 & 44.1 & 44.0 & 59.5 & \bf{34.5} & 66.6 & 53.4 & 52.7 & 82.1 & 52.7 & 9.4 & 89.3 & 50.5 & 38.5 & 53.8 \\
					SePiCo$\dag$~\cite{SePiCo} & 93.2 & 68.1 & 73.7 & 32.8 & 16.3 & 54.6 & \bf{49.5} & 48.1 & \bf{74.2} & 31.0 & \bf{86.3} & 57.9 & 50.9 & 82.4 & 52.2 & 1.3 & 83.8 & 43.9 & 29.8 & 54.2 \\
					MIC~\cite{MIC} & 88.2 & 60.5 & 73.5 & \bf{53.5} & 23.8 & 52.3 & 44.6 & 43.8 & 68.6 & 34.0 & 58.1 & 57.8 & 48.2 & 78.7 & 58.0 & \bf{13.3} & \bf{91.2} & 46.1 & \bf{42.9} & 54.6 \\
					\midrule
					Baseline &\bf{94.3} & \bf{70.0} & 77.4 & 40.8 & 13.8 & 53.3 & 28.9 & 44.7 & 66.4 & 34.1 & 81.4 & 57.1 & 42.7 & 81.3 & 49.6 & 5.0 & 89.4 & 50.5 & 35.8 & 53.5  \\
					Base.+MGCDA & 93.7 & 68.7 & 76.8 & 40.1 & 26.1 & \bf{56.9} & 49.0 & \bf{55.3} & 37.9 & 30.2 & 20.8 & \bf{59.3} & 49.6 & \bf{83.9} & 28.9 & 4.3 & 85.0 & \bf{52.3} & 34.1 & 50.2 \\
					CMDA($I$) & 93.4 & 65.6 & 76.0 & 40.9 & 22.4 & 54.8 & 48.5 & 47.6 & 65.7 & 30.2 & 78.1 & 56.8 & 46.9 & 80.8 & 64.2 & 12.9 & 74.7 & 44.5 & 37.0 & \bf{54.8} \\
					\bottomrule
				\end{tabular}
			\end{small}
		\end{center}
		\caption{Quantitative semantic segmentation results evaluated with MIoU (\%) in the image-based Dark Zurich Dataset. The best result is highlighted in bold.}
		\label{tab:STOA_DZ}
	\end{table*}
	
	\subsection{Datasets}
	\textbf{DSEC Night-Semantic Dataset.} To provide a benchmark for nighttime image-event semantic segmentation, we introduce the first image-event nighttime semantic segmentation dataset, \textit{i.e.,} DSEC Night-Semantic, based on the DSEC dataset~\cite{DSEC}. In DSEC, images and events are acquired by two different sensors which makes the two modalities not completely aligned. To obtain paired image-event data, we utilize depth data to warp from the image coordinates to the event coordinates with a resolution of 640$\times$480. Our dataset consists of 5 nighttime sequences of Zurich City 09a-e, and includes 1,692 training samples and 150 testing samples. For each testing sample, we manually annotate them in 18 classes: Road, Sidewalk, Building, Wall, Fence, Pole, Traffic Light, Traffic Sign, Vegetation, Terrain, Sky, Person, Rider, Car, Bus, Train, Motorcycle and Bicycle. 
	\revise{Detailed dataset split is shown in Table~\ref{tab:dataset}. Distribution of annotations across individual classes is provided in the supplemental material.}
	
	\textbf{Dark Zurich Dataset.} To thoroughly evaluate the effectiveness of our Image Content-Extractor, we conduct experiments on the image-based Dark Zurich dataset~\cite{DarkZurich}. Since there is no event modality in this dataset, we exclude $E$ along with steps 2 and 4 of Algorithm~\ref{alg:CMDA} during training.
	
	
	\begin{table}[t]
		\begin{center}
			\begin{tabular}{cccc}
				\toprule
				Method & MIoU($E$) & MIoU($I$) & MIoU($I$+$E$) \\
				\midrule
				Baseline & - & 55.06 & - \\
				Base. w/ $I_{CE}$ & - & 56.78 & - \\
				Base. w/ $E_{ME}$ & 45.06 & 53.46 & 55.65 \\
				CMDA & \bf{46.02} & \bf{57.76} & \bf{60.05} \\
				\bottomrule
			\end{tabular}
		\end{center}
		\caption{Ablation of $I_{CE}$ and $E_{ME}$ in our CMDA.}
		\label{tab:ablation}
	\end{table}

	\subsection{Comparison of SOTA Approaches}
	\textbf{DSEC Night-Semantic Dataset.} First, we compare our proposed CMDA with previous SOTA image-based unsupervised nighttime semantic segmentation approaches, including SePiCo~\cite{SePiCo}, Refign~\cite{Refign}, MIC~\cite{MIC}, and DAFormer~\cite{DAFormer}. The results in Table~\ref{tab:STOA} and Figure~\ref{fig:STOA} demonstrate the superior performance of our proposed CMDA, outperforming DAFormer~\cite{DAFormer} by +4.1\%. The fusion of high dynamic range event modality facilitates robust feature extracting from the scene, achieving improved nighttime semantic segmentation of 60.1\%.
	In addition, we find that training with the event modality and testing without it is also instrumental. The performance of CMDA($I$) is significantly improved compared to the baseline (+2.7\%), which indicates that events can guide the network in extracting more reliable features from images at night. Qualitative results in Figure~\ref{fig:STOA} demonstrate the substantial improvement in the segmentation of low-light objects and backgrounds. 
	
	\textbf{Dark Zurich Dataset.} In Table~\ref{tab:STOA_DZ}, we conduct experiments on the image-based Dark Zurich dataset to verify the effectiveness of our proposed Image Content-Extractor. First, we combine the day-to-night style transfer network of MGCDA~\cite{SSNight_MGCDA} with our baseline, and style transfer on the input domain is supposed to help the self-training framework in DAFormer~\cite{DAFormer} to alleviate the domain adaptation difficulties. However, the result is degraded (-3.3\%) due to the unrealistic and unreliable transferred images. In contrast, our proposed Image Content-Extractor eliminates most of the style information while preserving the content information, which surpasses the baseline by +1.3\% and achieves the SOTA MIoU score of 54.8\%.
	
	\subsection{Ablation Studies}
	Image Content-Extractor and Image Motion-Extractor are key components of the CMDA framework, bridging the gaps between domains and modalities. Table~\ref{tab:ablation} provides an overview of the ablation studies of these two components. 
	(1) The application of $I_{CE}$ results in an improvement of the baseline performance MIoU($I$) by +1.72\%, demonstrating the assistance of $I_{CE}$ for minimizing the domain shifts between the representations of daytime and nighttime images. (2) However, introducing event modality with only $E_{ME}$ impairs the features extraction of image. MIoU($I$) has a reduction of -1.6\% compared to the baseline and MIoU($I$+$E$) only has a minor improvement of +0.6\%. 
	We consider that when calculating $\mathcal{L}_{t}$, pseudo labels $\hat{Y}_t$ are generated by the fusion of both modalities. However, this fusion is unreliable at the beginning and hinders the initial training of the network, which in turn has a detrimental effect.
	(3) When employing both $I_{CE}$ and $E_{ME}$, we fuse $I$ and $E$/$I_{CE}$ randomly at each training step, which alleviates the above problem. The performance is further improved to 60.05\% MIoU($I$+$E$), improving +4.99\% compared to the baseline. More detailed ablation studies of the Image Motion-Extractor and Image Content-Extractor are shown below.
	
	\begin{figure}
		\centering
		\centerline{\includegraphics[scale=0.4]{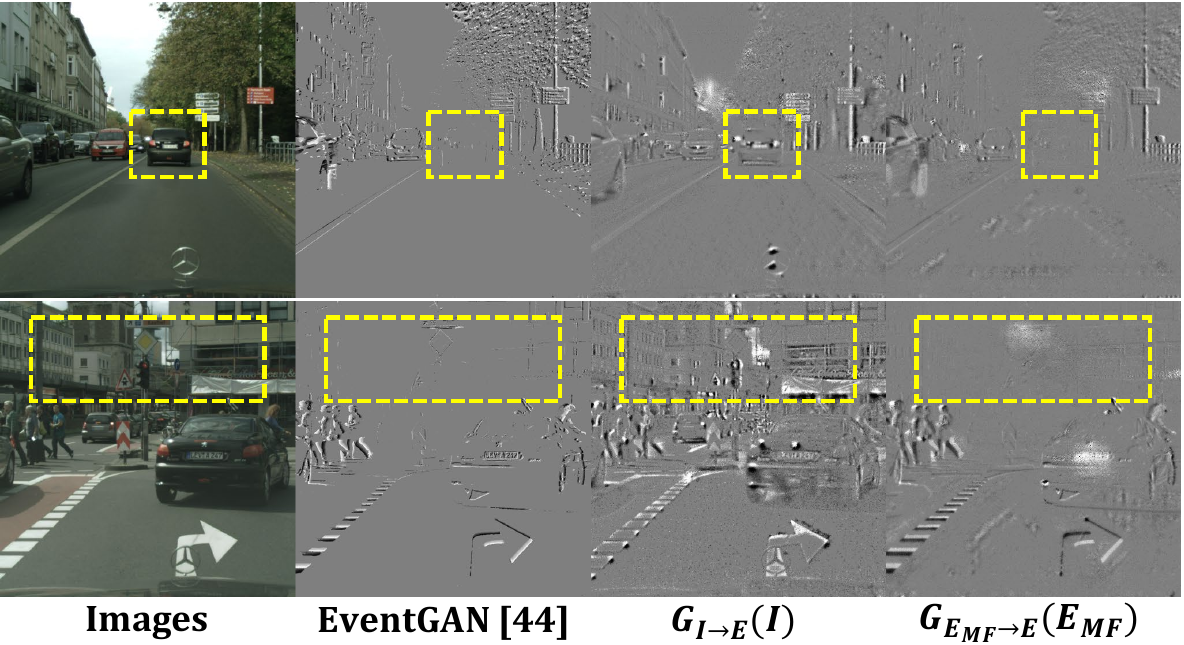}}
		\caption{Comparison of different ways to generate $\hat{E}_s$. As shown in the yellow box, $\hat{E}_s$ generated from a single image $G_{I\to E}(I)$ cannot simulate motion-related regions, which has a significant distribution difference from real events. In addition, $\hat{E}_s$ from EventGAN~\cite{EventGAN} does not construct the nighttime style.}
		\label{fig:events-generation}
	\end{figure}
	
	\begin{table}[t]
		\begin{center}
			\begin{tabular}{@{}c@{\enspace}c@{\enspace}c@{\enspace}c@{}}
				\toprule
				Method & MIoU($E$) & MIoU($I$) & MIoU($I$+$E$) \\
				\midrule
				\revise{ESIM\cite{ESIM} $\to E_t$} & 42.09 & 53.59 & 54.10 (+0.51) \\
				$I \to E_t$          & 41.81 & 54.21 & 54.50 (+0.29) \\
				$E_{ME} \to E_t$     & 44.91 & 55.47 & 56.63 (+1.16) \\
				EventGAN\cite{EventGAN} $\to E_t$ & 43.68 & 55.79 & 56.74 (+0.95) \\
				$I + G\to E_t$       & 39.03 & 55.24 & 57.21 (+1.97) \\
				$E_{ME} + G\to E_t$  & \bf{46.02} & \bf{57.76} & \bf{60.05} (+\bf{2.29}) \\
				\bottomrule
			\end{tabular}
		\end{center}
		\caption{Different approaches of adapting to nighttime event modality. The values in parentheses of MIoU($I$+$E$) represent the gain compared to MIoU($I$) after fusion with the event modality.}
		\label{tab:ablation_events}
	\end{table}
	
	\subsection{Image Motion-Extractor}
	We compare our Image Motion-Extractor with \revise{ESIM~\cite{ESIM} and EventGAN~\cite{EventGAN} that directly generate events from two temporally adjacent images}, and a straightforward approach that generates events from daytime images by a style transfer network $G$. Results are presented in Table~\ref{tab:ablation_events} and Figure~\ref{fig:events-generation}. 
	
	As demonstrated in Table~\ref{tab:ablation_events}, our proposed $E_{ME}$ exhibits superior MIoU($E$) performance compared to \revise{ESIM~\cite{ESIM} (+2.82\%) and} EventGAN~\cite{EventGAN} (+1.23\%), even when implemented without $G$. When combined with $G$, the proposed $E_{ME}+G$ achieves a remarkable improvement of +2.29\%. It surpasses the improvement +1.97\% of $I+G$ and achieves the SOTA performance of 60.05\%.
	
	Visualization of $\hat{E}_s$ is shown in Figure~\ref{fig:events-generation}. EventGAN~\cite{EventGAN} ignores the noise of event cameras at night, and $\hat{E}_s$ generated by $I$ depicts all edges in the scene, which fails to accurately simulate the motion-capture property of event cameras. By employing $E_{ME}$ with $G$, our $\hat{E}_s$ simulates events only in the regions with the relative motion and achieves a more accurate depiction of nighttime events. 
	
	\begin{figure}[t]
		\centering
		\centerline{\includegraphics[scale=0.82]{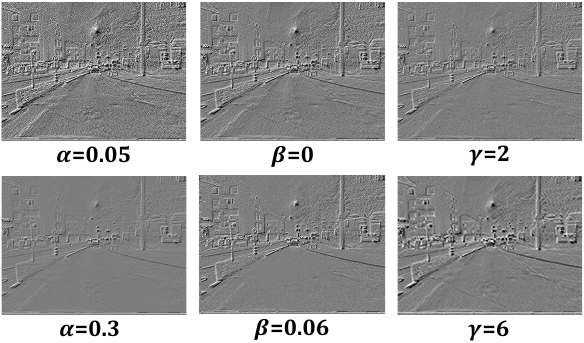}}
		\caption{Visualization of nighttime $I_{CE}$ generated with different parameters.}
		\label{fig:isd}
	\end{figure}
	
	\subsection{Image Content-Extractor} 
	Our Image Content-Extractor plays a key role in bridging the domain gap between daytime and nighttime images. In Figure~\ref{fig:isd}, we provide a visualization of nighttime $I_{CE}$ generated with $\alpha$, $\beta$ in Eq.~\ref{eqn:ClipIgn} and $\gamma$ in Eq.~\ref{eqn:ISF}. 
	$\alpha$ controls the lower-bound and upper-bound of $\operatorname{F_{LogDiff}}(I_1, I_2)$. A large value of $\alpha$ narrows down the effective information in the scene, while a small value of $\alpha$ amplifies the proportion of noise.
	$\beta$ aims to filter out the values less than $\beta$. A smaller $\beta$ will retain more noise while a larger $\beta$ will destroy the information of the scene.
	$\gamma$ controls the shift pixels of the image relative to itself. A small value of $\gamma$ can better capture scene details. Conversely, a large value of $\gamma$ blurs the edges. Experiments in Table~\ref{tab:alpha_gamma} demonstrate that the moderate values of $\alpha$, $\beta$, and small value of $\gamma$ have the optimal trade-off. 

	\begin{table}[t]
		\begin{center}
			\begin{tabular}{ccccc}
				\toprule
				$\alpha$ & 0.05   & \colorbox{gray!25}{0.1}   & 0.15   & 0.2    \\
				\midrule
				MIoU($I$+$E$)   & 57.59 & \bf{60.05} & 59.43 & 59.70 \\
				\midrule
				\midrule
				$\beta$ & 0      & \colorbox{gray!25}{0.005} & 0.015  & 0.03     \\
				\midrule
				MIoU($I$+$E$)  & 58.38 & \bf{60.05} & 59.04 & 57.61 \\
				\midrule
				\midrule
				$\gamma$ & \colorbox{gray!25}{1}     & 2     & 3     & 4     \\
				\midrule
				MIoU($I$+$E$)  & \bf{60.05} & 59.40 & 59.28 & 58.57 \\
				\bottomrule
			\end{tabular}
		\end{center}
		\caption{Analysis of $\alpha$, $\beta$ and $\gamma$. When adjusting one parameter, the other two parameters in the gray background remain unchanged.}
		\label{tab:alpha_gamma}
	\end{table}

	\section{Conclusion}
	We introduce a novel framework, Cross-Modality Domain Adaptation (CMDA), \revise{for semantic segmentation on} nighttime image and event modalities. Our proposed Image Motion-Extractor and Image Content-Extractor effectively bridge the gaps between modalities and domains. Notably to the best of our knowledge, our work is the first to introduce event modality into nighttime semantic segmentation. To facilitate our research, we present the DSEC Night-Semantic dataset that comprises 1,692 training samples and 150 testing samples. \revise{A comprehensive evaluation demonstrates} that our CMDA achieves substantial performance improvements and effectively leverages \revise{the complementary modalities}.
	
	\revise{\textbf{Acknowledgment.} This work was supported in part by the National Key Research and Development Program of China under Grant 2021YFB1714300, in part by the National Natural Science Foundation of China under Grant 62233005, in part by the Program of Shanghai Academic Research Leader under Grant 20XD1401300, in part by the Sino-German Center for Research Promotion under Grant M-0066, and in part by the Program of Introducing Talents of Discipline to Universities through the 111 Project under Grant B17017.}

	{\small
		\bibliographystyle{ieee_fullname}
		\bibliography{egbib}
	}

%% file: mysecondfile.tex
		
		\title{Supplementary Material---CMDA: Cross-Modality Domain Adaptation for Nighttime Semantic Segmentation}

		\twocolumn[{
			
			\author{Ruihao Xia$^{1}$~~~~Chaoqiang Zhao$^{1}$~~~~Meng Zheng$^{2}$~~~~Ziyan Wu$^{2}$~~~~Qiyu Sun$^{1}$~~~~Yang Tang$^{1*}$
			\and
			$^{1}$East China University of Science and Technology~~~~$^{2}$United Imaging Intelligence
			\and
			{\tt\small \{xia\_rho,~zhaocq,~qysun\}@mail.ecust.edu.cn,~\{meng.zheng,~ziyan.wu\}@uii-ai.com}
			\and
			{\tt\small yangtang@ecust.edu.cn}
		}
		
		\maketitle
		
		\vspace{-1cm}
		\begin{center}
			\renewcommand{\tabcolsep}{1pt}
			\includegraphics[scale=0.69]{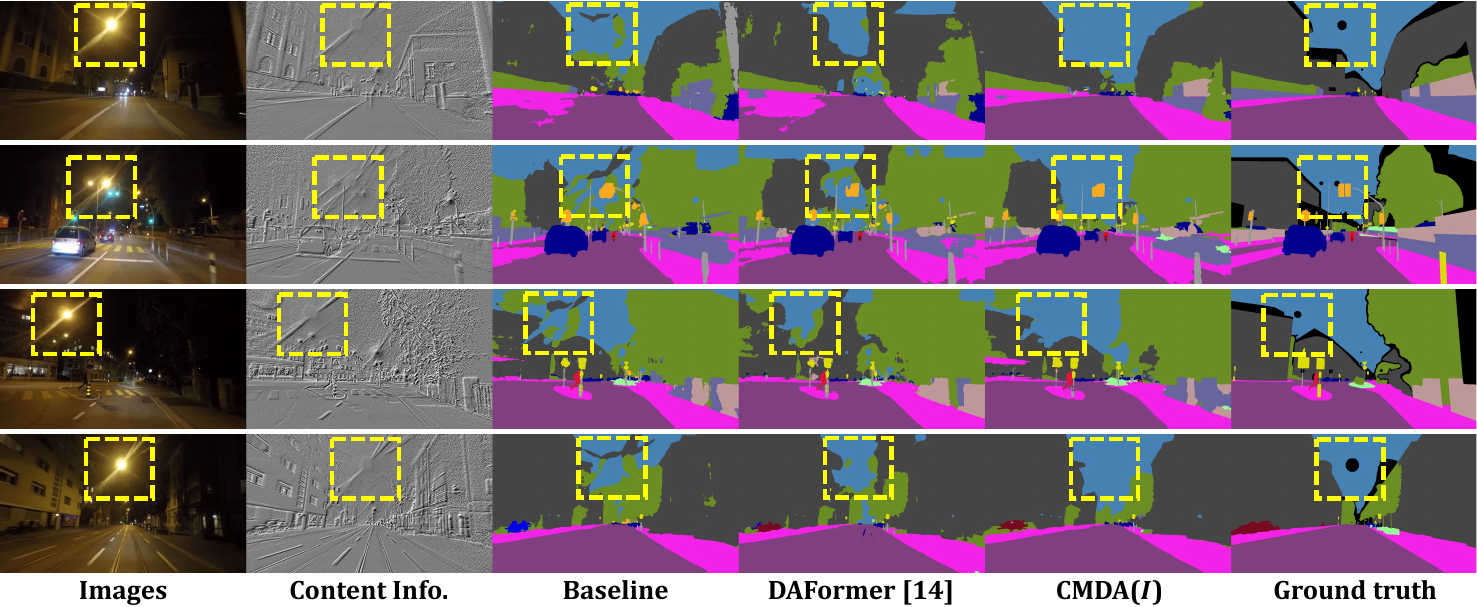}
			\label{fig:dz}
		\end{center}
		\vspace{-0.2cm}
		\hypertarget{fig:dz}{Figure 1. }{Qualitative results of our baseline, SOTA approach DAFormer~\cite{DAFormer}, and our proposed CMDA($I$) in the image-based Dark Zurich dataset~\cite{DarkZurich}. Note that the content information generated by our proposed Image Content-Extractor are only utilized during training in CMDA($I$).}
		\vspace{0.5cm}
	}]
	
	\setcounter{figure}{1}
	
	\setcounter{section}{0}
	\section{Event Representation}
	\blfootnote{$^{*}$Corresponding author.}
	The event camera outputs a continuous stream of events, wherein each event consists of four distinct elements, namely $(t, x, y, p)$. Here, $t$ denotes the trigger time, $(x,y)$ represents the spatial coordinate, and $p \in \{+1,-1\}$ is the polarity that represents the sign of the brightness change~\cite{EventCameras2}. 
	
	Raw events are discrete spatial-temporal points that pose challenges for feature extraction and integration with image modalities. To overcome this, we follow the previous approach~\cite{VoxelGrid} to embed raw events as an image $E\in \mathbb{R} ^{H\times W\times B}$, where $B$ represents the number of temporal bins. A higher value of $B$ indicates a more refined representation of temporal information. However, in our proposed CMDA, we focus on the High Dynamic Range (HDR) of the event camera instead of the high temporal resolution. Moreover, to ensure consistency in the number of channels of $E_{ME}$ and $E$ for training the style transfer network $G_{E_{ME}\to E}$, we set $B=1$.
	
	\section{Annotations Distribution}
	\revise{Our proposed DSEC Night-Semantic dataset contains 18 classes. Distribution of annotations across individual classes is provided in Figure~\ref{fig:dataset}.}
	
	\begin{figure*}[t]
		\centering
		\includegraphics[scale=0.68]{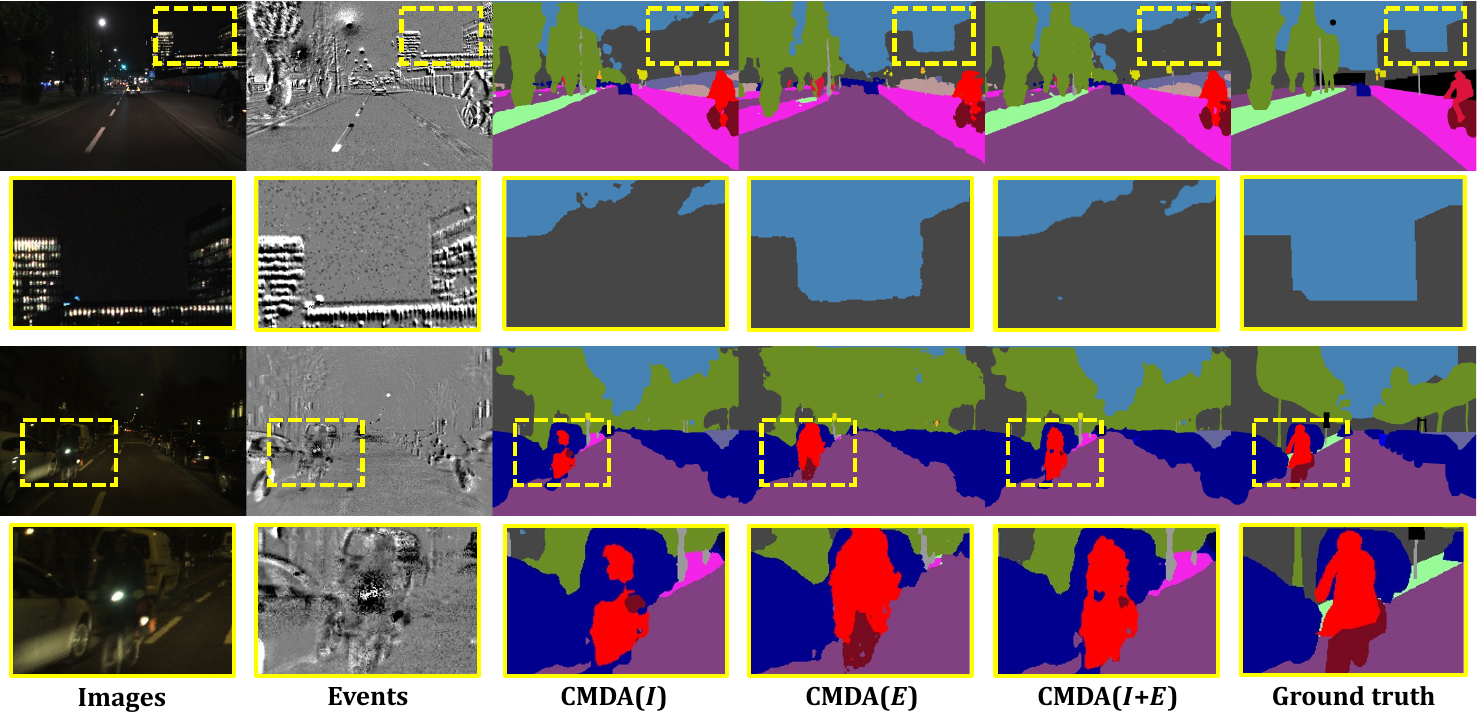}
		\caption{The failure cases of our CMDA in the proposed DSEC Night-Semantic image-event dataset.}
		\label{fig:FailedCases}
	\end{figure*}
	
	\begin{figure}
		\centering
		\includegraphics[scale=0.53]{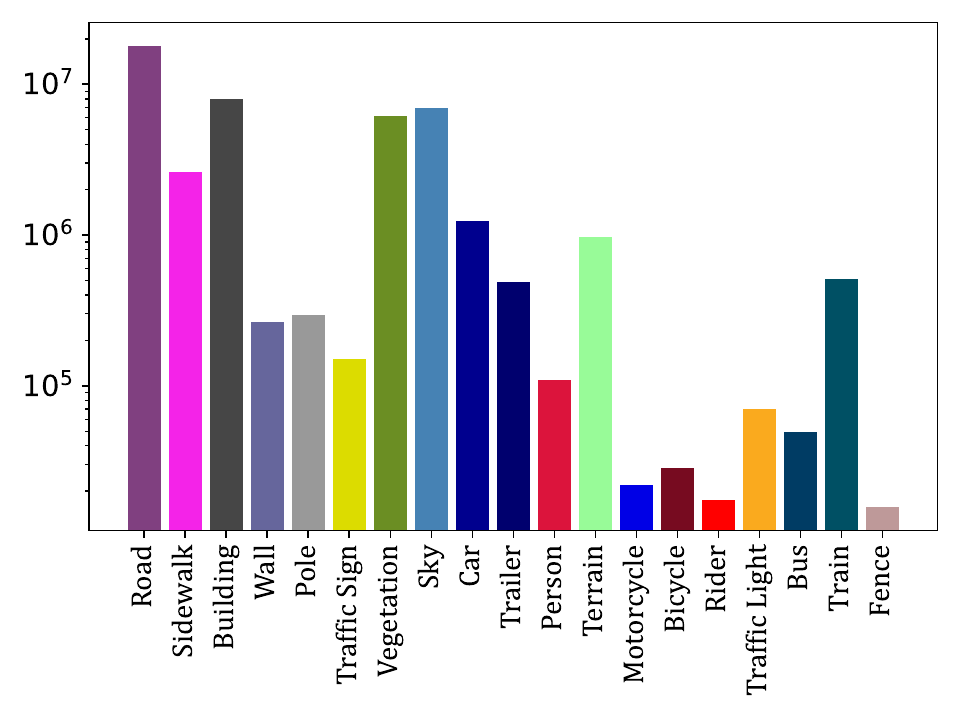}
		\caption{\revise{Number of annotated pixels (y-axis) per classes (x-axis) for our proposed DSEC Night-Semantic dataset.}}
		\label{fig:dataset}
	\end{figure}
	
	
	\section{Training details}
	\revise{\textbf{Style Transfer Network $G_{E_{ME}\to E}$.} Following CycleGAN~\cite{CycleGAN}, we randomly select 1,000 $E_{ME}$ and $E_{t}$ from the Cityscapes and DSEC dataset. Then, cycle consistency and adversarial loss are utilized to train the network for 200 epochs.}
	
	\textbf{Data Augmentation.} In the source domain, namely the Cityscapes~\cite{Cityscapes} dataset, we resize $I_s$, $\hat{E}_s$, and $I_{CE\_s}$ to $1024\times512$ and randomly crop them into $512\times512$, as per the DAFormer~\cite{DAFormer}. 
	
	For the target domain, namely our proposed DSEC Night-Semantic dataset, we randomly crop areas of $400\times400$ on $I_t$, $E_t$, and $I_{CE\_t}$ and resize them to $512\times512$. 
	
	During the calculation of the target loss $\mathcal{L}_t$, we follow DACS~\cite{DACS} and apply additional data augmentation techniques, \textit{i.e.,} color jitter, Gaussian blur, and ClassMix~\cite{ClassMix}, on the input images $I_t$ of $f^S$. The corresponding $I_{CE\_t}$ are directly generated from $I_t$ with the proposed Image-Content Extractor, while $E_t$ are exclusively enhanced by ClassMix~\cite{ClassMix}.

	\section{Visualization in Dark Zurich}
	In this section, we demonstrate the performance of the proposed Image Content-Extractor in the image-based Dark Zurich dataset~\cite{DarkZurich}, and compare it with the SOTA approach DAFormer~\cite{DAFormer}. As shown in Figure~\hyperlink{fig:dz}{1}, our proposed Image Content-Extractor effectively mitigates the impact of nighttime glare, resulting in clearer edge segmentation of the sky and other objects.
	
	

	\section{Failure Cases}
	Our proposed CMDA integrates the event modality into nighttime semantic segmentation for the first time, leading to a significant improvement in segmentation performance. However, our CMDA may fail to generate satisfactory results in some cases. We compare these results with different modalities inputs in Figure~\ref{fig:FailedCases}. 
	
	Looking at the event modality in the first row, it is evident that the HDR of nighttime events provides a clear contrast between the edges of buildings. Consequently, the building and the sky in the yellow box of CMDA($E$) are accurately segmented with event inputs. However, when fusing images with events, CMDA($I+E$) failed to fully utilize the benefits of the event modality. The results in the second row shows a similar situation, where events capture more robust features in the corner cases, yet CMDA fails to integrate events effectively.
	
	The aforementioned cases show that our CMDA puts higher weights on image modality during fusion, which results in the under-utilization of  event modality. We attribute this to the fact that in the source domain, daytime images typically contain a vast majority of favorable information in the scene. As a result, CMDA can generate satisfactory segmentation results even when just relying on the image modality, and the weights of the event modality is lowered. Conversely, in nighttime scenes, event modality demonstrates its HDR advantage. Nonetheless, due to the absence of ground truth for supervised training, pseudo labels generated by $f^T$ tend to rely more on the image modality.
	
	\section{Limitations}
	\begin{itemize}
		\item \textbf{Paired Images and Events.} Our CMDA requires nighttime paired event and image modalities for training, so we wrap the $1440\times1080$ images to the $640\times480$ event coordinates. However, this operation compromises the advantage of high-resolution in the original images, and may have a negative impact on fine-grained segmentation. Therefore, future studies could focus on how to directly fuse unpaired image and event modalities, thereby leveraging the high resolution of images and HDR nighttime events.
		\item \textbf{Short-Time Events.} Our CMDA employs events captured within 50ms window as input. However, it is worth noting that short-time events may not provide a comprehensive representation of the scene, particularly when the relative motion between the scene and the event camera is weak. This is precisely why we choose to fuse events with images rather than relying solely on events. Therefore, future studies could explore approaches to express a comprehensive representation of the scene by utilizing events over an extended time range.
		\item \textbf{Reliability of Generated Events $\hat{E}_s$.} \revise{The ESIM events simulator~\cite{ESIM} guarantees the high temporal resolution of events by interpolating a large number of frames between two adjacent images. Conversely, our proposed Image Motion-Extractor utilizes only the difference between two images to simulate events, which undoubtedly ignores the temporal information of the events and generates unreliable events compared to real events. However, in this paper, we focus mainly on the high dynamic range advantages provided by the event modality. Thus, events are embedded as a single channel image form and the temporal information is discarded. Future studies could explore the impact of high temporal resolution of events on nighttime semantic segmentation.}
		
	\end{itemize}
